\documentclass[conference]{IEEEtran}
\IEEEoverridecommandlockouts
\ifCLASSOPTIONcompsoc
  \usepackage[nocompress]{cite}
\else
  \usepackage{cite}
\fi

\usepackage{booktabs}
\usepackage{graphicx}
\usepackage{multirow}
\usepackage{amsmath}
\usepackage{amssymb}
\usepackage{makecell}
\usepackage[table]{xcolor}
\definecolor{bestblue}{RGB}{147,197,253}
\definecolor{secondblue}{RGB}{219,234,254}
\usepackage{hyperref}
\hypersetup{
    colorlinks=true,
    linkcolor=blue,
    urlcolor=magenta,
    citecolor=blue
}

\begin{document}
\title{Seeing Time: Visual-Temporal Representation Learning for Interpretable Time Series Clustering}

\author{
Zheng Zhu\textsuperscript{1}, Zexi Tan\textsuperscript{2,$*$},
Yuming Deng\textsuperscript{2} and Yiqun Zhang\textsuperscript{3}\\[1ex]
\textsuperscript{1}Huizhou No. 1 High School (High School Student),
\textsuperscript{2}Guangdong University of Technology \\
\textsuperscript{3}South China University of Technology\\
zhuz3398199086@foxmail.com,
\{tanzexi, dengyuming\}@mails.gdut.edu.cn,
yqzhangzyq@gmail.com
\thanks{$*$ Corresponding author}
         
}
\maketitle

\begin{abstract}
Multivariate Time Series (MTS) clustering is an important tool in temporal data mining, aiming to discover latent group structures from complex observations without supervision.
Although existing deep clustering methods can learn discriminative temporal representations, the resulting latent clusters are often difficult to relate back to waveform characteristics that practitioners can directly inspect and compare, limiting their ability to assess whether the discovered patterns reflect meaningful temporal behaviors.
This paper, therefore, proposes WAVE (Waveform Aligned Visual-temporal Embedding), which treats time series and their deterministically rendered waveform plots as complementary views of the same observations.
To produce discriminative representations whose cluster structures can be traced to observable waveform characteristics, WAVE aligns and integrates fine-grained temporal variations with holistic visual patterns, while associating each discovered cluster with its centroid-nearest authentic sample. Accordingly, interpretability in this work specifically refers to waveform-level traceability rather than a general explanation of model decisions.
Extensive evaluations across 10 real-world public datasets show that WAVE achieves the highest macro-averaged clustering performance and the best average rank among the compared methods, while qualitative case studies illustrate how the discovered clusters can be inspected through authentic waveform records.
The source code is available at \url{https://github.com/Zheng-Zhu1/WAVE}.
\end{abstract}


\IEEEpeerreviewmaketitle

\section{Introduction}
Multivariate Time Series (MTS) clustering \cite{TanAAAI2026} is widely
applied in industrial monitoring~\cite{wang2023d3r}, medical diagnosis~\cite{cai2024jolt},
human activity recognition~\cite{dhekane2025transfer}, and environmental sensing~\cite{liang2023airformer}, aiming
to discover latent groups and representative patterns from
complex records when expert annotations are
scarce or costly to obtain. In recent years, deep clustering methods
have substantially improved pattern discovery in complex data
by learning discriminative temporal representations~\cite{draayer2025deep}.
However, these methods primarily rely on time series for representation learning~\cite{WANG2026112899}, making
the resulting abstract cluster structures difficult to relate
to waveform characteristics that can be intuitively observed
and compared~\cite{icdm2024nasl}. Consequently, users struggle to determine
how different clusters vary in terms of overall contours,
peak-valley distributions, and periodic morphology, and also
find it difficult to validate the clustering results
against the original time series records~\cite{icdm2024rankscl}. Therefore, how to make the discovered cluster structures traceable to observable
waveform characteristics while preserving clustering
discriminability remains an urgent MTS clustering problem~\cite{WangTFMCC2026}. In this work, interpretability specifically refers to
waveform-level traceability: the ability to associate discovered
clusters with authentic waveform records that can be directly
inspected and compared.

Recent methods primarily improve the discriminability of MTS cluster structures by encoding local variations, long-term dependencies, and cross-channel relationships from raw sequences through reconstruction~\cite{he2026gtm}, contrastive learning~\cite{liu2024timesurl}, or clustering objectives~\cite{Fu_Hu_2025}. 
Yet discriminability alone does not make the discovered structures traceable to observable temporal patterns. 
This limitation motivates two related directions:model interpretation~\cite{xie2026kdd} and result visualization~\cite{zhang2024ssl4ts}. Nevertheless, their optimization processes and final clustering remain confined to the feature space, and the resulting clusters can usually be evaluated only through clustering metrics or dimensionality-reduced distributions~\cite{LI2025MVCIMTS}. 
Regarding model interpretation, some studies employ saliency scores~\cite{moebus2025contimask} or prototype representations~\cite{nguyen2024robust} to identify local factors that influence model outputs, but they still struggle to understand and compare different clusters in terms of waveform characteristics such as overall contours, peak-valley distributions, and periodic morphology. 
As for result visualization, waveform plots are typically used only to display clustering results and do not participate in representation learning or the formation of cluster structures~\cite{TKDE2025Boniol}. 
Therefore, existing studies still lack a unified approach that directly incorporates observable waveform characteristics into representation learning and establishes connections among sequence data, cluster structures, and real waveforms~\cite{piao2024fredformer}.

\begin{figure*}[!t]
\centering
\includegraphics[width=\textwidth]{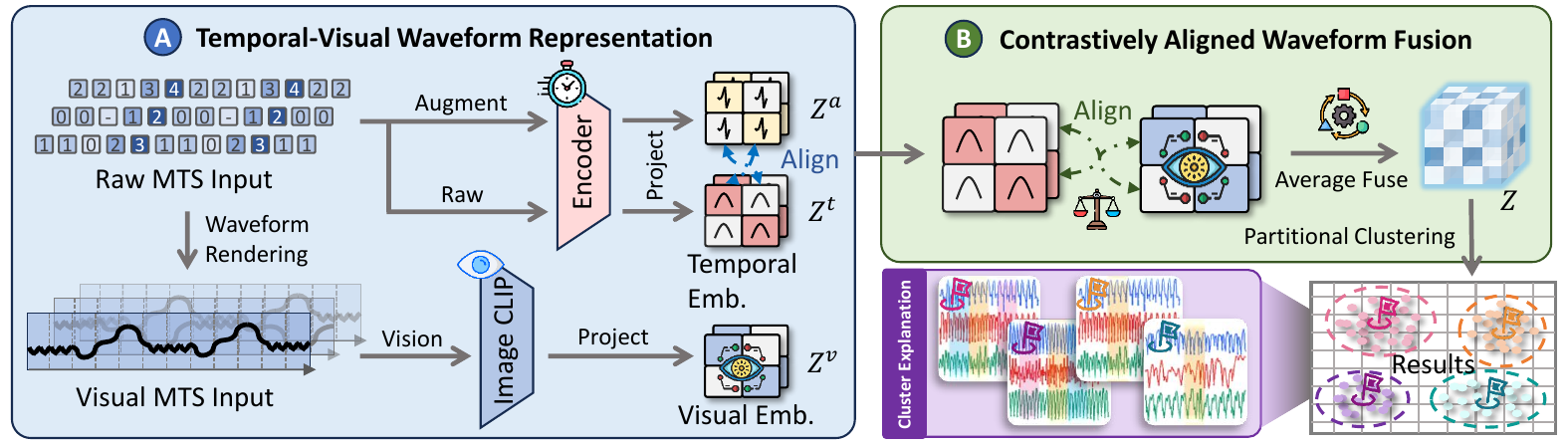}
\caption{Overview of WAVE. (A) Each MTS record, its weakly augmented sequence, and its deterministically rendered waveform image are encoded into temporal, augmented, and visual representations $\mathbf{Z}^{t}$, $\mathbf{Z}^{a}$, and $\mathbf{Z}^{v}$. (B) Cross-modal contrastive alignment maps the temporal and visual representations into a shared space, followed by parameter-free normalized equal fusion to obtain $\mathbf{Z}$. The fused representations are clustered, and the real sample nearest to each centroid is selected as an observable cluster representative.}
\label{fig:overview}
\end{figure*}

This gap raises a more fundamental question: whether waveform morphology can participate in clustering representation learning rather than remain merely an external basis for interpreting the results~\cite{icdm2024rankscl}. 
Waveform plots explicitly expose temporal characteristics such as global contours, peak-valley distributions, and periodic morphology~\cite{gao2024units}, providing observable structural evidence complementary to sequence representations.
Meanwhile, recent studies have shown that transforming structured non-visual data into images allows pretrained visual models to capture holistic patterns that are difficult to characterize solely from raw sequences~\cite{zhong2025timevlm}. 
These findings reveal a potential complementarity: sequence encoders preserve fine-grained temporal variations, whereas visual encoders provide a more holistic description of waveform morphology. 
Realizing this complementarity is nontrivial because the two encoders organize information in distinct representational geometries~\cite{kimura2025infomae}, and the informativeness of their representations may vary across samples~\cite{mohapatra2025maestro}. 
Simply combining them cannot guarantee a coherent clustering representation. 
Thus, the central challenge is to incorporate observable waveform characteristics into cluster formation while establishing correspondence between the two representational spaces and preserving their respective strengths~\cite{nam2025bimodal}. 

According to this observation, we propose WAVE (Waveform Aligned Visual-temporal Embedding), a visual–temporal framework for observable MTS clustering. 
To preserve fine-grained measurements while incorporating waveform morphology, WAVE represents each record as a sequence and a multichannel waveform plot, capturing temporal variations and holistic patterns, respectively. 
Because the two views are encoded independently, their shared origin does not guarantee representational consistency. We therefore employ cross-modal contrastive learning to align the paired views, while temporal augmentation improves representation stability. Furthermore, the aligned representations are integrated through parameter-free normalized equal fusion.
Finally, WAVE selects the record nearest to each cluster center as its representative waveform, making abstract clusters directly observable and comparable.
The main contributions are summarized below.

\begin{itemize}
    \item We propose WAVE, a visual-temporal MTS clustering framework that addresses the limited interpretability of existing methods primarily optimized for clustering performance by incorporating observable waveform morphology directly into representation learning and cluster formation.

    \item WAVE grounds each learned cluster in an authentic waveform by selecting the real record nearest to its centroid. This association makes cluster structures observable in the original data space and supports direct comparison among discovered groups.

    \item By coupling temporal discrimination with observable waveform characteristics, WAVE produces accurate and interpretable cluster structures. Experiments on 10 real-world datasets verify its clustering effectiveness and the visual traceability of the discovered groups.
\end{itemize}

\section{Proposed Method}

Given an unlabeled MTS dataset
$\mathcal{X}=\{\mathbf{X}_i\}_{i=1}^{N}$, where
$\mathbf{X}_i\in\mathbb{R}^{T\times C}$ contains $C$ variables over $T$ time steps, WAVE learns a representation for clustering, as illustrated in Fig.~\ref{fig:overview}. 
An MTS record contains both fine-grained temporal dynamics and global waveform morphology, which may not be captured equally well by a single encoder. Therefore, WAVE represents each record through two paired views: the original sequence and its annotation-free waveform image. The former preserves sequential dependencies, while the latter makes geometric patterns such as peaks, slopes, and periodicity directly observable.

\subsection{Temporal-Visual Waveform Representation}

Sequence-only representations can capture temporal dynamics, but their latent cluster structures are difficult to relate to observable waveform characteristics. To preserve discriminative temporal information while grounding the learned clusters in directly inspectable patterns, WAVE constructs two complementary views of each record. The sequence view retains temporal ordering and value variations, while the waveform view exposes structural morphology. They are encoded separately and subsequently aligned in a shared space.

For the temporal branch, discriminative patterns may range from short local changes to dependencies spanning the entire sequence. Each variable is first normalized independently to prevent scale differences from dominating the representation. Multi-scale one-dimensional convolutions then capture local patterns at different temporal ranges, and a Transformer encoder models their long-range relationships. Attention pooling summarizes globally informative regions, while max pooling preserves salient local responses. This design produces a representation sensitive to both distributed temporal context and brief but distinctive events:
\begin{equation}
\mathbf{z}_i^{t}
=
\operatorname{norm}\left(
p_t\left(f_t(\mathbf{X}_i)\right)
\right)
\in\mathbb{R}^{d},
\end{equation}
where $f_t(\cdot)$ and $p_t(\cdot)$ denote the temporal encoder and projection head, respectively.

However, the temporal embedding remains latent and does not directly expose the waveform morphology underlying the learned clusters. WAVE therefore renders each normalized record into an annotation-free waveform image $\mathbf{I}_i=\mathcal{R}(\mathbf{X}_i)$ using a fixed rendering scheme, such that visual differences mainly reflect waveform structure. A frozen pretrained OpenCLIP ViT-B/32 image encoder extracts morphological patterns, and a trainable projection head maps them into the representation space:
\begin{equation}
\mathbf{z}_i^{v}
=
\operatorname{norm}\left(
p_v\left(f_v(\mathbf{I}_i)\right)
\right).
\end{equation}
Although both temporal and visual views describe the same record, their embeddings are produced by heterogeneous encoders and therefore remain geometrically inconsistent. Thus, explicit cross-modal alignment is required before complementary signatures can be reliably fused for clustering.

\subsection{Contrastively Aligned Waveform Fusion}

To establish the required cross-view correspondence, WAVE treats $(\mathbf{z}_i^{t},\mathbf{z}_i^{v})$ from the same record as a positive pair and cross-record pairs as negatives. A symmetric cross-modal contrastive objective increases the similarity of matched pairs while separating mismatched ones, encouraging each record to preserve its identity across the two modalities. Consequently, proximity in the aligned space reflects agreement between fine-grained temporal dynamics and observable waveform morphology, providing a consistent basis for their subsequent fusion.

The aligned normalized embeddings are then combined with equal coefficients:
\begin{equation}
\mathbf{z}_i
=
\operatorname{norm}\left(
\mathbf{z}_i^{t}
+
\mathbf{z}_i^{v}
\right).
\end{equation}
This parameter-free fusion allows the clustering space to reflect both fine-grained temporal dynamics and observable waveform morphology. Consequently, waveform structure participates directly in determining sample proximity and cluster formation.

\subsection{Optimization and Clustering}

Training couples the temporal-visual alignment established above with temporal consistency under weak augmentation. The overall objective is
\begin{equation}
\mathcal{L}
=
\mathcal{L}_{\mathrm{tv}}
+
\lambda\mathcal{L}_{\mathrm{ta}},
\end{equation}
where
$\mathcal{L}_{\mathrm{tv}}
=\mathcal{L}_{\mathrm{NCE}}(\mathbf{Z}^{t},\mathbf{Z}^{v})$
establishes temporal-visual correspondence, while
$\mathcal{L}_{\mathrm{ta}}
=\mathcal{L}_{\mathrm{NCE}}(\mathbf{Z}^{t},\mathbf{Z}^{a})$
constrains weakly augmented sequences to remain close to their original representations.
The coefficient $\lambda$ controls the relative strength of $\mathcal{L}_{\mathrm{ta}}$.

To specify these contrastive objectives, let $\alpha,\beta\in\{t,v,a\}$ index the temporal, visual, and augmented representations, and let $i,j\in\{1,\ldots,B\}$ index samples within a batch. The directional loss is
\begin{equation}
\ell_{\alpha\rightarrow\beta}^{(i)}
=
-\log
\frac{
\exp\!\left(s(\mathbf{z}_i^{\alpha},\mathbf{z}_i^{\beta})/\tau\right)
}{
\sum_{j=1}^{B}
\exp\!\left(s(\mathbf{z}_i^{\alpha},\mathbf{z}_j^{\beta})/\tau\right)
},
\end{equation}
where $s(\cdot,\cdot)$ denotes cosine similarity and $\tau$ is the temperature. WAVE optimizes both directions through the symmetric objective
\begin{equation}
\mathcal{L}_{\mathrm{NCE}}
(\mathbf{Z}^{\alpha},\mathbf{Z}^{\beta})
=
\frac{1}{2B}
\sum_{i=1}^{B}
\left(
\ell_{\alpha\rightarrow\beta}^{(i)}
+
\ell_{\beta\rightarrow\alpha}^{(i)}
\right).
\end{equation}

After optimization, the fused representations defined in the preceding subsection are standardized and partitioned by minimizing the $k$-means objective
\begin{equation}
\min_{\{\boldsymbol{\mu}_k\},\{q_i\}}
\sum_{i=1}^{N}
\left\|
\widetilde{\mathbf{z}}_i-\boldsymbol{\mu}_{q_i}
\right\|_2^2,
\end{equation}
where $\widetilde{\mathbf{z}}_i$ is the standardized fused representation, $q_i$ is its cluster assignment, and $\boldsymbol{\mu}_k$ denotes the centroid of $k$-th cluster.

To connect each latent cluster back to an observable waveform, WAVE selects the authentic sample closest to its centroid as the cluster exemplar:
\begin{equation}
e_k
=
\arg\min_{i:q_i=k}
\left\|
\widetilde{\mathbf{z}}_i-\boldsymbol{\mu}_k
\right\|_2.
\end{equation}
The corresponding original record $\mathbf{X}_{e_k}$ provides a directly inspectable waveform for the discovered cluster.

\section{Experiments}

This paper evaluates WAVE through clustering, ablation, and interpretability experiments on ten UEA MTS datasets~\cite{bagnall2018uea}: AtrialFibrillation, BasicMotions, Cricket, ERing, Epilepsy, HandMovementDirection, Libras, NATOPS, RacketSports, and StandWalkJump. Following the transductive protocol, training and test sets are combined for unsupervised learning, with labels used only for evaluation and the cluster number set to the ground-truth class count. 
The baselines cover both recent clustering-specific approaches, including EMTC~\cite{TanAAAI2026}, TFMCC~\cite{WangTFMCC2026}, FCACC~\cite{WANG2026112899}, MVCIMTS~\cite{LI2025MVCIMTS}, and $k$-Graph~\cite{TKDE2025Boniol}, and general time series representation learners, including GTM~\cite{he2026gtm}, FEI~\cite{Fu_Hu_2025}, TimesURL~\cite{liu2024timesurl}, and UNITS~\cite{gao2024units}. Performance is measured by ACC, NMI, and ARI. 
Ablations examine the two representation views and training objectives, while the interpretability study evaluates waveform exemplars associated with the discovered clusters. We report mean and standard deviation over five seeds. WAVE is implemented in PyTorch 2.8.0 and trained for up to 100 epochs on an RTX 5090 with batch size 32 and $\lambda=0.3$. The temporal encoder contains six Transformer layers and eight heads, OpenCLIP ViT-B/32 remains frozen, and the embedding dimension is 256.

\subsection{Clustering Performance Evaluation}

Table~\ref{tab:aggregate_results} summarizes performance across
all ten datasets. WAVE achieves the highest macro-averaged ACC,
NMI, and ARI and the best average rank on all three metrics.
It ranks first in 15 and within the top two in 23 of the 30
dataset-metric combinations, with particularly clear gains on
Epilepsy and Libras. Friedman tests confirm overall differences
among the methods for all metrics ($p<0.001$), although pairwise advantages over the strongest baseline do not remain significant after Holm correction. These results demonstrate improved average performance while also indicating dataset-dependent benefits. Complete results are provided in Table~I of the
\href{https://github.com/Zheng-Zhu1/WAVE/blob/main/ICDM-Appendix.pdf}{appendix}.

\begin{table}[t]
\centering
\caption{Aggregate clustering results on ten UEA datasets.
Avg.\ and Rank denote macro-average performance and average rank,
respectively. Lower ranks are better; dark and light blue indicate
the best and second-best results.}
\label{tab:aggregate_results}

\scriptsize
\setlength{\tabcolsep}{1.8pt}
\renewcommand{\arraystretch}{1.10}

\begin{tabular*}{\columnwidth}
{@{\extracolsep{\fill}}lcccccc@{}}
\toprule

\multirow{2}{*}{\textbf{Methods}}
& \multicolumn{2}{c}{\textbf{ACC}}
& \multicolumn{2}{c}{\textbf{NMI}}
& \multicolumn{2}{c}{\textbf{ARI}}
\\

\cmidrule(lr){2-3}
\cmidrule(lr){4-5}
\cmidrule(lr){6-7}

& \textbf{Avg.} & \textbf{Rank}
& \textbf{Avg.} & \textbf{Rank}
& \textbf{Avg.} & \textbf{Rank}
\\

\midrule

TimesURL (AAAI'24)
& 0.499 & 5.100
& 0.395 & 4.950
& 0.280 & 4.900
\\

UNITS (NeurIPS'24)
& 0.418 & 7.300
& 0.239 & 7.700
& 0.134 & 7.700
\\

FEI (AAAI'25)
& 0.511 & 5.050
& 0.362 & 5.300
& 0.252 & 5.250
\\

MVCIMTS (INFFUS'25)
& 0.291 & 9.450
& 0.116 & 8.050
& 0.060 & 8.200
\\

k-Graph (TKDE'25)
& 0.480 & 6.050
& 0.310 & 6.400
& 0.208 & 6.500
\\

EMTC (AAAI'26)
& 0.544 & \cellcolor{secondblue}3.500
& 0.374 & 4.600
& 0.230 & 4.800
\\

TFMCC (AAAI'26)
& \cellcolor{secondblue}0.578 & 4.150
& \cellcolor{secondblue}0.453
& \cellcolor{secondblue}4.050
& \cellcolor{secondblue}0.347
& \cellcolor{secondblue}3.700
\\

FCACC (PR'26)
& 0.531 & 5.950
& 0.400 & 4.500
& 0.292 & 5.200
\\

GTM (ICLR'26)
& 0.448 & 6.300
& 0.262 & 7.350
& 0.156 & 6.600
\\

\textbf{WAVE (Ours)}
& \cellcolor{bestblue}\textbf{0.703}
& \cellcolor{bestblue}\textbf{2.150}
& \cellcolor{bestblue}\textbf{0.587}
& \cellcolor{bestblue}\textbf{2.100}
& \cellcolor{bestblue}\textbf{0.520}
& \cellcolor{bestblue}\textbf{2.150}
\\

\bottomrule
\end{tabular*}
\end{table}

\begin{table}[t]
\centering
\caption{Ablation results averaged over ten UEA datasets.
Drop denotes Full WAVE minus the corresponding variant.
$^{*}$ indicates a significant degradation according to a
two-sided Wilcoxon signed-rank test with Holm correction
($p<0.05$). Complete results are provided in the Appendix.}
\label{tab:ablation_average}

\scriptsize
\setlength{\tabcolsep}{1.5pt}
\renewcommand{\arraystretch}{1.10}

\begin{tabular*}{\columnwidth}
{@{\extracolsep{\fill}}lcccccc@{}}
\toprule

\multirow{2}{*}{\textbf{Variant}}
& \multicolumn{2}{c}{\textbf{ACC}}
& \multicolumn{2}{c}{\textbf{NMI}}
& \multicolumn{2}{c}{\textbf{ARI}}
\\

\cmidrule(lr){2-3}
\cmidrule(lr){4-5}
\cmidrule(lr){6-7}

& \textbf{Avg.} & \textbf{Drop}
& \textbf{Avg.} & \textbf{Drop}
& \textbf{Avg.} & \textbf{Drop}
\\

\midrule

Full WAVE
& \textbf{0.703} & --
& \textbf{0.587} & --
& \textbf{0.520} & --
\\

\midrule

w/o Visual
& 0.663 & 0.040
& 0.543 & 0.044
& 0.469 & 0.052
\\

w/o Temporal
& 0.666 & $0.037^{*}$
& 0.519 & $0.068^{*}$
& 0.451 & $0.069^{*}$
\\

\midrule

w/o $\mathcal{L}_{\mathrm{tv}}$
& 0.581 & $\mathbf{0.121}^{*}$
& 0.415 & $\mathbf{0.172}^{*}$
& 0.346 & $\mathbf{0.174}^{*}$
\\

w/o $\mathcal{L}_{\mathrm{ta}}$
& 0.687 & 0.016
& 0.569 & $0.018^{*}$
& 0.498 & $0.022^{*}$
\\

\bottomrule
\end{tabular*}
\end{table}

\subsection{Ablation Study}

Table~\ref{tab:ablation_average} summarizes the view and objective
ablations over all ten datasets. Removing
$\mathcal{L}_{\mathrm{tv}}$ causes the largest and most consistent
degradation, reducing ACC, NMI, and ARI by 0.121, 0.172, and 0.174,
respectively. Removing the temporal view also produces significant
reductions across all three metrics, confirming its importance for
preserving fine-grained temporal dynamics. In contrast, removing
$\mathcal{L}_{\mathrm{ta}}$ leads to smaller drops that are
significant for NMI and ARI but not ACC, indicating a mainly
regularizing role. Removing the visual view decreases all three
macro-averaged metrics, although the differences are not
statistically significant after Holm correction, suggesting
dataset-dependent contributions. Overall, these results identify
explicit temporal-visual alignment as WAVE's most consistent
mechanism. Complete dataset-level results are reported in
Table~II of the
\href{https://github.com/Zheng-Zhu1/WAVE/blob/main/ICDM-Appendix.pdf}
{appendix}.

Beyond these component ablations, Table~\ref{tab:visual_encoder}
examines different visual encoders under the same ten-dataset
protocol. Pretrained OpenCLIP substantially outperforms its randomly
initialized counterpart, confirming the benefit of visual
pretraining for capturing waveform morphology. ResNet18 achieves
competitive but slightly lower macro-averaged performance,
indicating that the gains arise from pretrained visual
representations rather than being exclusive to CLIP.

\begin{table}[t]
\centering
\caption{Macro-averaged clustering performance of different visual
encoders on ten UEA datasets over five seeds.}
\label{tab:visual_encoder}
\scriptsize
\renewcommand{\arraystretch}{1.08}
\begin{tabular*}{0.92\columnwidth}
{@{\extracolsep{\fill}}llccc@{}}
\toprule
\textbf{Encoder}
& \textbf{Pretraining}
& \textbf{ACC}
& \textbf{NMI}
& \textbf{ARI} \\
\midrule
OpenCLIP ViT-B/32
& LAION-2B
& \textbf{0.703}
& \textbf{0.587}
& \textbf{0.520} \\
OpenCLIP ViT-B/32
& Random
& 0.614
& 0.474
& 0.394 \\
ResNet18
& ImageNet-1K
& 0.686
& 0.567
& 0.504 \\
\bottomrule
\end{tabular*}
\end{table}

\begin{figure}
\centering
\includegraphics[width=\columnwidth]{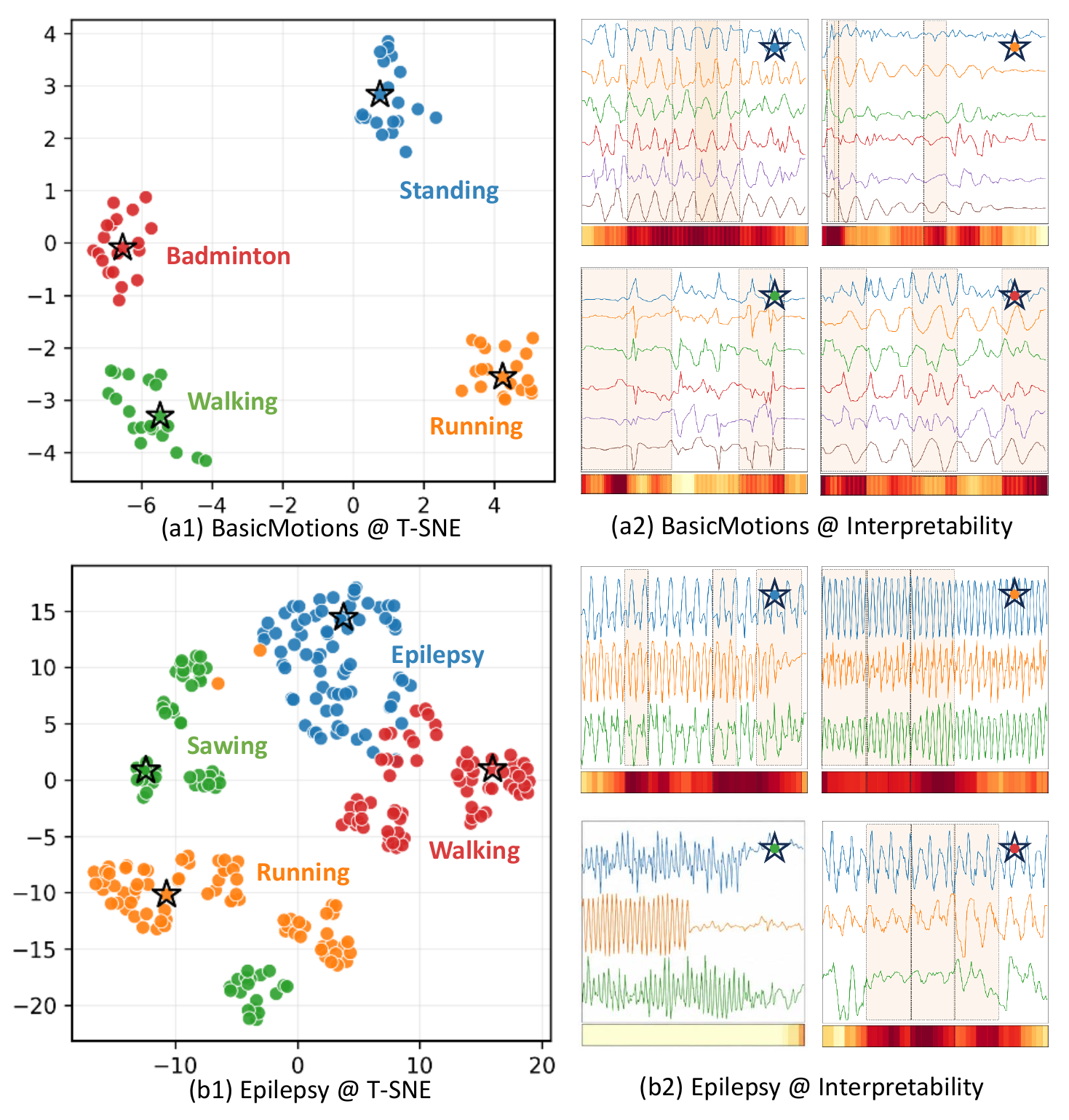}
\caption{Qualitative clustering and waveform-traceability results
on BasicMotions and Epilepsy. The left panels show t-SNE projections of the learned representations, where stars mark the authentic samples nearest to the cluster centroids. The right panels show the corresponding waveform records for direct morphological inspection.}
\label{interpret}
\end{figure}

\subsection{Interpretability Case Study}

Fig.~\ref{interpret} provides a case-level view of how WAVE's temporal-visual mechanism shapes the learned clusters on BasicMotions and Epilepsy. The temporal branch preserves activity-dependent dynamics, while the waveform view exposes complementary morphological cues such as periodicity, amplitude, and irregular variations. Through cross-view alignment, samples sharing these temporal and visual characteristics are drawn into consistent regions of the embedding space. The centroid-nearest representatives further show that the learned clusters correspond to distinctive waveform structures. These observations are consistent with WAVE's mechanism of organizing MTS through complementary temporal dynamics and visual morphology.

\begin{figure}[t]
\centering
\includegraphics[width=0.98\columnwidth]{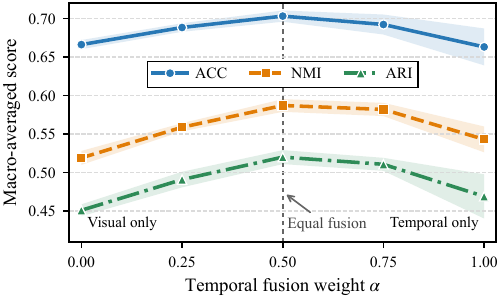}
\caption{Sensitivity to the temporal fusion weight $\alpha$ over
ten UEA datasets. The visual weight is $1-\alpha$. Curves report
macro-averaged performance, and shaded regions indicate one
standard deviation across five seeds.}
\label{fig:fusion_sensitivity}
\end{figure}

\subsection{Fusion-Weight Sensitivity}

Fig.~\ref{fig:fusion_sensitivity} evaluates the fixed fusion $z(\alpha)=\alpha\,\mathrm{norm}(z^{t})+(1-\alpha)\,\mathrm{norm}(z^{v})$.
Intermediate weights outperform both single-view endpoints across all three metrics. Equal fusion achieves the highest ACC and ARI and performs comparably to $\alpha=0.75$ in NMI. The narrower uncertainty bands around intermediate weights also indicate greater stability across random seeds. The stable performance around $\alpha=0.5$ supports equal fusion as a robust parameter-free
choice without dataset-specific tuning.

\section{Concluding Remarks}

MTS clustering remains challenging because cluster structure may arise from both temporal dynamics and waveform morphology, while either view alone can be incomplete. WAVE aligns and fuses these complementary views into a unified representation and links each cluster to an authentic waveform sample for
inspection. Results on 10 UEA datasets, together with ablation and sensitivity analyses, support its clustering effectiveness and the importance of cross-view alignment. WAVE currently assumes fixed waveform rendering and known cluster numbers, leaving adaptive rendering and cluster-number estimation for future work.

\section*{Acknowledgements}

This work was supported in part by the National Natural Science Foundation of China (NSFC) and the Guangdong Provincial Special Fund for Science and Technology Innovation Strategy. Zheng Zhu, the high-school student first author, led development, implementation, evaluation, result analysis, and manuscript drafting. Zexi Tan and Yuming Deng provided methodological guidance and result verification. Yiqun Zhang supervised the project, coordinated administration, secured funding, and revised the manuscript.

\bibliographystyle{IEEEtran}
\bibliography{mybib}

@inproceedings{icdm2024nasl,
  author    = {Duy A. Nguyen and others},
  title     = {Improving Time Series Encoding with Noise-Aware Self-Supervised Learning and an Efficient Encoder},
  booktitle = {ICDM},
  pages     = {340--349},
  year      = {2024},
  doi       = {10.1109/ICDM59182.2024.00041}
}

@inproceedings{icdm2024rankscl,
  author    = {Qianying Ren and others},
  title     = {Rank Supervised Contrastive Learning for Time Series Classification},
  booktitle = {ICDM},
  pages     = {839--844},
  year      = {2024},
  doi       = {10.1109/ICDM59182.2024.00102}
}

@article{dhekane2025transfer,
  author    = {Sourish Gunesh Dhekane and others},
  title     = {Transfer Learning in Sensor-Based Human Activity Recognition: A Survey},
  journal   = {ACM Comput. Surv.},
  volume    = {57},
  number    = {8},
  articleno = {205},
  pages     = {1--39},
  year      = {2025},
  doi       = {10.1145/3717608}
}

@article{draayer2025deep,
  author    = {Erick Draayer and others},
  title     = {Deep Clustering for Large-Scale Interpretable Time Series Segmentation},
  journal   = {Data Min. Knowl. Discov.},
  volume    = {40},
  number    = {1},
  pages     = {1--36},
  year      = {2025},
  doi       = {10.1007/s10618-025-01170-y}
}

@article{LI2025MVCIMTS,
  title     = {Contrastive learning-based multi-view clustering for incomplete multivariate time series},
  journal   = {Inf. Fusion},
  volume    = {117},
  pages     = {102812},
  year      = {2025},
  author    = {Yurui Li and others}
}

@article{nguyen2024robust,
  author    = {Thu Trang Nguyen and others},
  title     = {Robust Explainer Recommendation for Time Series Classification},
  journal   = {Data Min. Knowl. Discov.},
  volume    = {38},
  pages     = {3372--3413},
  year      = {2024},
  doi       = {10.1007/s10618-024-01045-8}
}

@article{TKDE2025Boniol,
  author    = {Paul Boniol and others},
  title     = {$k$-Graph: A Graph Embedding for Interpretable Time Series Clustering},
  year      = {2025},
  volume    = {37},
  number    = {5},
  journal   = {IEEE Trans. Knowl. Data Eng.},
  pages     = {2680--2694}
}

@article{WANG2026112899,
  title     = {Fuzzy cluster-aware contrastive clustering for time series},
  journal   = {Pattern Recognit.},
  volume    = {173},
  pages     = {112899},
  year      = {2026},
  author    = {Congyu Wang and others}
}

@article{zhang2024ssl4ts,
  author    = {Kexin Zhang and others},
  title     = {Self-Supervised Learning for Time Series Analysis: Taxonomy, Progress, and Prospects},
  journal   = {IEEE Trans. Pattern Anal. Mach. Intell.},
  volume    = {46},
  number    = {10},
  pages     = {6775--6794},
  year      = {2024},
  doi       = {10.1109/TPAMI.2024.3387317}
}

@inproceedings{cai2024jolt,
  author    = {Yifu Cai and others},
  title     = {{JoLT}: Jointly Learned Representations of Language and Time-Series for Clinical Time-Series Interpretation},
  booktitle = {AAAI},
  volume    = {38},
  number    = {21},
  pages     = {23447--23448},
  year      = {2024},
  doi       = {10.1609/aaai.v38i21.30423}
}

@inproceedings{Fu_Hu_2025,
  title     = {Frequency-Masked Embedding Inference: A Non-Contrastive Approach for Time Series Representation Learning},
  booktitle = {AAAI},
  year      = {2025},
  author    = {En Fu and Yanyan Hu},
  pages     = {16639--16647}
}

@inproceedings{gao2024units,
  author    = {Shanghua Gao and others},
  title     = {{UniTS}: A Unified Multi-Task Time Series Model},
  booktitle = {NeurIPS},
  volume    = {37},
  pages     = {140589--140631},
  year      = {2024}
}

@inproceedings{he2026gtm,
  title     = {{GTM}: A General Time-series Model for Enhanced Representation Learning of Time-Series data},
  author    = {Cheng He and others},
  booktitle = {ICLR},
  year      = {2026},
  url       = {https://openreview.net/forum?id=PWM6FERWz9}
}

@inproceedings{kimura2025infomae,
  author    = {Tomoyoshi Kimura and others},
  title     = {{InfoMAE}: Pair-Efficient Cross-Modal Alignment for Multimodal Time-Series Sensing Signals},
  booktitle = {WWW},
  pages     = {3084--3095},
  year      = {2025},
  doi       = {10.1145/3696410.3714853}
}

@inproceedings{liang2023airformer,
  author    = {Yuxuan Liang and others},
  title     = {{AirFormer}: Predicting Nationwide Air Quality in {China} with Transformers},
  booktitle = {AAAI},
  volume    = {37},
  number    = {12},
  pages     = {14329--14337},
  year      = {2023},
  doi       = {10.1609/aaai.v37i12.26676}
}

@inproceedings{liu2024timesurl,
  author    = {Jiexi Liu and others},
  title     = {{TimesURL}: Self-Supervised Contrastive Learning for Universal Time Series Representation Learning},
  booktitle = {AAAI},
  volume    = {38},
  number    = {12},
  pages     = {13918--13926},
  year      = {2024},
  doi       = {10.1609/aaai.v38i12.29299}
}

@inproceedings{moebus2025contimask,
  author    = {Max Moebus and others},
  title     = {Contimask: Explaining Irregular Time Series via Perturbations in Continuous Time},
  booktitle = {NeurIPS},
  pages     = {1--26},
  year      = {2025}
}

@inproceedings{mohapatra2025maestro,
  author    = {Payal Mohapatra and others},
  title     = {{MAESTRO}: Adaptive Sparse Attention and Robust Learning for Multimodal Dynamic Time Series},
  booktitle = {NeurIPS},
  pages     = {1--32},
  year      = {2025}
}

@inproceedings{nam2025bimodal,
  author    = {Youngeun Nam and others},
  title     = {Bi-Modal Learning for Networked Time Series},
  booktitle = {KDD},
  volume    = {2},
  pages     = {2162--2173},
  year      = {2025},
  doi       = {10.1145/3711896.3736856}
}

@inproceedings{piao2024fredformer,
  author    = {Xihao Piao and others},
  title     = {Fredformer: Frequency Debiased Transformer for Time Series Forecasting},
  booktitle = {KDD},
  pages     = {2400--2410},
  year      = {2024},
  doi       = {10.1145/3637528.3671928}
}

@inproceedings{TanAAAI2026,
  title     = {Mask the Redundancy: Evolving Masking Representation Learning for Multivariate Time-Series Clustering},
  volume    = {40},
  doi       = {10.1609/aaai.v40i30.39777},
  number    = {30},
  booktitle = {AAAI},
  author    = {Zexi Tan and others},
  year      = {2026},
  pages     = {25787--25795}
}

@inproceedings{wang2023d3r,
  author    = {Chengsen Wang and others},
  title     = {Drift Doesn't Matter: Dynamic Decomposition with Diffusion Reconstruction for Unstable Multivariate Time Series Anomaly Detection},
  booktitle = {NeurIPS},
  volume    = {36},
  pages     = {10758--10774},
  year      = {2023}
}

@inproceedings{WangTFMCC2026,
  title     = {Time-Frequency Augmented Multi-level Contrastive Clustering for Time Series},
  booktitle = {AAAI},
  author    = {Congyu Wang and others},
  year      = {2026},
  pages     = {26142--26150}
}

@inproceedings{zhong2025timevlm,
  author    = {Siru Zhong and others},
  title     = {{Time-VLM}: Exploring Multimodal Vision-Language Models for Augmented Time Series Forecasting},
  booktitle = {ICML},
  pages     = {78478--78497},
  year      = {2025}
}

@inproceedings{xie2026kdd,
    author = {Tao Xie and others},
    title = {AnchorMoE: Interpretable Time Series Classification via Anchor-Routed MoE},
    year = {2026},
    booktitle = {KDD},
    pages = {5720--5731}
}

@misc{bagnall2018uea,
      title={The UEA Multivariate Time Series Classification Archive, 2018}, 
      author={Anthony Bagnall and others},
      year={2018},
      howpublished  = {arXiv preprint arXiv:1811.00075}
}

\end{document}